\pdfoutput=1

\PassOptionsToPackage{table}{xcolor}

\documentclass[11pt]{article}

\usepackage[]{acl}

\usepackage{times}
\usepackage{latexsym}

\usepackage[T1]{fontenc}

\usepackage[utf8]{inputenc}

\usepackage{microtype}

\usepackage{inconsolata}

\usepackage[table]{xcolor}
\usepackage{graphicx}
\usepackage{floatrow}
\usepackage{subcaption}
\usepackage{multirow}
\usepackage{amssymb}
\usepackage[flushleft]{threeparttable}
\usepackage{amsmath}
\usepackage{booktabs}

\usepackage{lipsum}
\usepackage{pifont}
\definecolor{applegreen}{rgb}{0.0, 0.5, 0.0}
\usepackage[normalem]{ulem}
\newcommand\sbullet[1][.5]{\mathbin{\vcenter{\hbox{\scalebox{#1}{$\bullet$}}}}}
\usepackage{longtable}
\newcommand*{\matryoshka}{%
    \raisebox{-.3\baselineskip}{%
        \includegraphics[
        height=\baselineskip,
        width=\baselineskip,
        keepaspectratio,
        ]{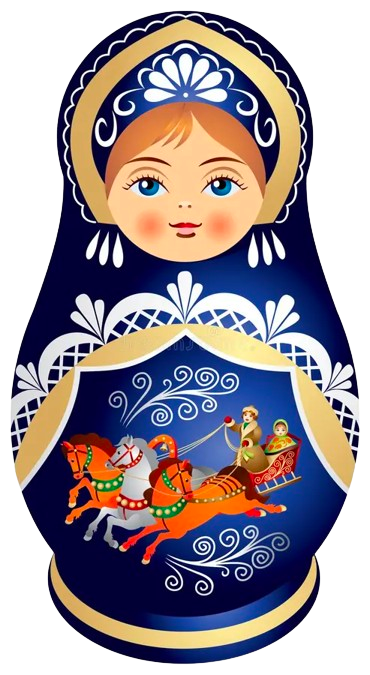}%
    }%
}
\newcommand*{\teachericon}{%
    \raisebox{-.3\baselineskip}{%
        \includegraphics[
        height=\baselineskip,
        width=\baselineskip,
        keepaspectratio,
        ]{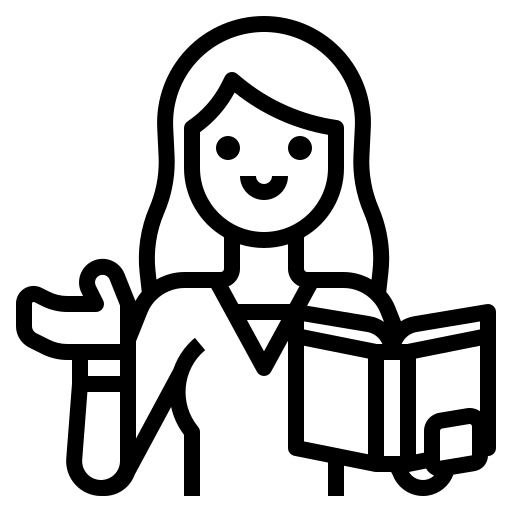}%
    }%
}

\newcommand*{\studenticon}{%
    \raisebox{-.3\baselineskip}{%
        \includegraphics[
        height=\baselineskip,
        width=\baselineskip,
        keepaspectratio,
        ]{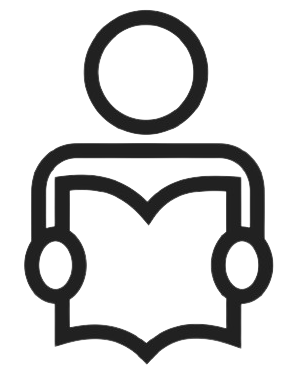}%
    }%
}

\title{2D Matryoshka Sentence Embeddings\thanks{Preprint. Work in progress.}}

\author{
    Xianming Li \includegraphics[width=0.35cm]{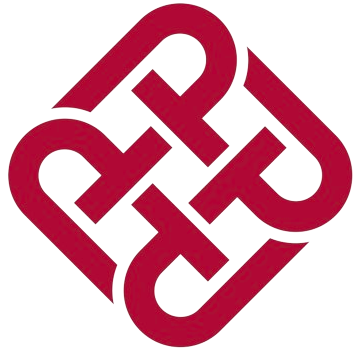},
    Zongxi Li \includegraphics[width=0.32cm]{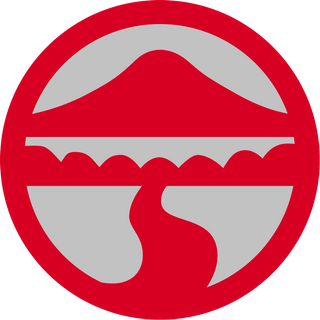} ,
    Jing Li \includegraphics[width=0.35cm]{figures/polyu.png} \thanks{\ \ Corresponding author},
    Haoran Xie \includegraphics[width=0.32cm]{figures/lingnan.png} ,
    Qing Li \includegraphics[width=0.35cm]{figures/polyu.png}\\
    \includegraphics[width=0.35cm]{figures/polyu.png} Department of Computing, The Hong Kong Polytechnic University, Hong Kong SAR \\
    \includegraphics[width=0.32cm]{figures/lingnan.png} Department of Computing and Decision Sciences, Lingnan University, Hong Kong SAR\\
  \texttt{xianming.li@connect.polyu.hk}\\
  \texttt{jing-amelia.li@polyu.edu.hk}\\
 }

\begin{document}  
\maketitle
\begin{abstract}

Common approaches rely on fixed-length embedding vectors from language models as sentence embeddings for downstream tasks such as semantic textual similarity (STS). Such methods are limited in their flexibility due to unknown computational constraints and budgets across various applications. Matryoshka Representation Learning (MRL) \cite{aditya2022matryoshka} encodes information at finer granularities, i.e., with lower embedding dimensions, to adaptively accommodate \emph{ad hoc} tasks. Similar accuracy can be achieved with a smaller embedding size, leading to speedups in downstream tasks. Despite its improved efficiency, MRL still requires traversing all Transformer layers before obtaining the embedding, which remains the dominant factor in time and memory consumption. This prompts consideration of whether the fixed number of Transformer layers affects representation quality and whether using intermediate layers for sentence representation is feasible. In this paper, we introduce a novel sentence embedding model called \textit{Two-dimensional Matryoshka Sentence Embedding} (2DMSE)\footnote{Our code is available at \url{https://github.com/SeanLee97/AnglE/blob/main/README_2DMSE.md}.}. It supports elastic settings for both embedding sizes and Transformer layers, offering greater flexibility and efficiency than MRL. We conduct extensive experiments on STS tasks and downstream applications. The experimental results demonstrate the effectiveness of our proposed model in dynamically supporting different embedding sizes and Transformer layers, allowing it to be highly adaptable to various scenarios.

\end{abstract}

\section{Introduction}
\label{sec::intro}
Sentence embedding learning \citep{conneau-etal-2017-supervised,cer-etal-2018-universal,sbert-nils-2019,simcse_gao_2021,li2023angle} is a fundamental task in semantic textual similarity (STS). 
It captures essential semantic and syntactic information in language, playing a crucial role in various scenarios such as retrieval augmented generation \citep{gao2023retrieval} and semantic duplication removal \citep{li2024generative}. 
The conventional deployment pipeline consists of two steps: \textbf{(1)} the forward pass to compute the representation and \textbf{(2)} the utilization of representations in downstream tasks \cite{aditya2022matryoshka}. 

\begin{figure*}
    \centering
    \includegraphics[width=0.98\textwidth]{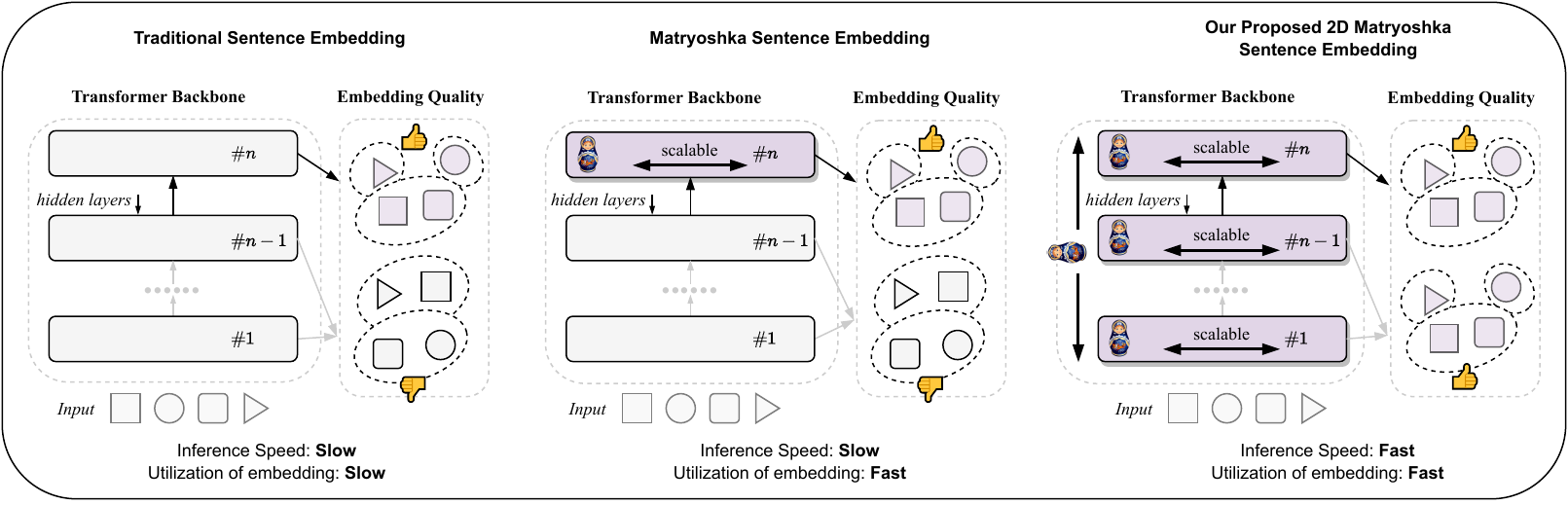}
    \caption{
    A visual comparison of various sentence embedding methods. The gray blocks represent Transformer layers fine-tuned with AnglE, which are not optimized for matryoshka representation. The purple block represents Transformer layers fine-tuned with AnglE together with matryoshka loss.
    }
    \label{figure_case}
\end{figure*}

Existing works \citep[\emph{inter alia}]{sbert-nils-2019,simcse_gao_2021,li2023angle} commonly select the last Transformer layer with full hidden size for all tasks, regardless of varying resources and requirements. However, \citet{aditya2022matryoshka} argues that using full-capacity embedding in such methods leads to unnecessary computational redundancy, as deep learning models tend to diffuse information, which could be encoded with fewer bits, across the high-dimensional vector. To inject elasticity and scalability into representation dimensions, \citet{aditya2022matryoshka} proposed $\mathrm{Matryoshka}$ $\mathrm{Representation}$ $\mathrm{Learning}$ (MRL). MRL derives information-rich low-dimensional vectors from the same high-dimensional representation in a nested fashion, resembling human perception of the natural world \cite{hegde2008time}. Given one pretrained language model, MRL yields a set of coarse-to-fine-grained representations while preserving main semantics. With up to a $14\times$ reduction in embedding dimensions, MRL achieves speedup in step \textbf{(2)} tasks such as classification and retrieval.

However, it is important to note that MRL is only scalable to the embedding of the last Transformer layer. Despite providing efficiency for downstream applications, MRL incurs an expensive and constant inference stage, i.e., step \textbf{(1)} for calculating full-throughput embedding vectors at all layers, as the forward-pass pipeline remains unchanged. This imposes a high computational requirement for deploying MRL, particularly when the encoding model is relatively large. Moreover, we examined STS performance using embedding vectors from shallow layers of BERT$_{base}$ finetuned by AnglE\footnote{We use BERT$_{base}$ as the base model in this work. For concise expression, we will use $\mathrm{AnglE}$ to denote the base model finetuned by AnglE. $\mathrm{MRL}$ and the proposed $\mathrm{2DMSE}$ also use AnglE for sentence embedding learning.} \citep{li2023angle}, the state-of-the-art sentence embedding learning method, and observed unexpected performance drops from intermediate layers. These observations inspired us to further study representation capacity from another dimension: the depth of Transformer layers, in addition to the embedding size.

In this paper, we introduce the Two-dimensional $\mathrm{Matryoshka}$ $\mathrm{Sentence}$ $\mathrm{Embedding}$ \matryoshka$^\mathbf{2}$ (2DMSE). Given a pretrained language model, 2DMSE aims to extend the original MRL's flexibility in sentence embedding learning by improving representation capacity at shallow layers. At each training step, we randomly sample a layer from the Transformer backbone (except for the last layer) following a uniform distribution. The proposed 2DMSE finetunes the last layer's embedding and the sampled layer's embedding simultaneously and in the matryoshka style for sentence embedding learning. Moreover, to further enhance the performance of shallow layers, we align their embeddings with those of the last layer for self-supervision, also following the matryoshka principle, by minimizing their Kullback-Leibler divergence. In this framework, shallow layers are explicitly involved in the representation learning process and are trained to become as powerful as the last layer. Therefore, every shallow layer is expected to be comparable with its subsequent layer, achieving a layer-level matryoshka effect through the continual pipeline. 
The key differences between the traditional sentence embedding approach, Matryoshka sentence embedding, and our proposed 2DMSE are depicted in Figure \ref{figure_case}.

2DMSE offers several advantages in sentence embedding learning. First, it significantly improves the performance of shallow layers' embeddings on STS benchmarks. The shallow layers' embeddings can already achieve acceptable performance, and substantial improvements are observed over the full-capacity embeddings, even without using additional training samples. Furthermore, the two-dimensional matryoshka training strategy makes the embedding model scalable and, most importantly, truncatable at two dimensions, i.e., the model depth and the embedding size, which can significantly enhance the efficiency and flexibility of utilizing 2DMSE embeddings. Given a large-scale language model, one can customize their Matryoshka models at different scales with a specified number of layers and embedding size according to the environment's requirements and computational resources of an \textit{ad hoc} task. Remarkably, the smaller models derived from 2DMSE can outperform their independently trained counterparts.

In summary, our contributions are as follows: 

$\bullet$ We propose the Two-dimensional $\mathrm{Matryoshka}$ $\mathrm{Sentence}$ $\mathrm{Embedding}$ \matryoshka$^\mathbf{2}$ (2DMSE) framework for flexible and scalable sentence embedding learning.

$\bullet$ 2DMSE supports elastic configurations for both model depth and embedding size with marginal overhead and seamlessly adapts to different deployment requirements.

$\bullet$ Extensive experiments suggest that 2DMSE outperforms powerful baselines and demonstrates excellent scalability.

\section{Related Work}
Our work focuses on embedding learning, specifically in the context of sentence embeddings. 
While early efforts focused primarily on word embeddings \citep{word2vec_mikolov_2013}, sentence embeddings allow for semantic representation with richer contextual information.
To better learn sentence embeddings, \textbf{supervised approaches} \citep{conneau-etal-2017-supervised,cer-etal-2018-universal,sbert-nils-2019,li2023angle} aligned with human supervision, thereby improving sentence embedding quality.
Recently, \textbf{contrastive learning} techniques \citep{carlsson2020semantic, zhang-etal-2020-unsupervised, giorgi-etal-2021-declutr, simcse_gao_2021, consert_yan_2021, chuang-etal-2022-diffcse, promcse_jiang_2022, zhuo-etal-2023-whitenedcse, xu-etal-2023-distillcse} were used to improve sentence embeddings further with in-batch negative learning.
With the advent of LLMs \citep{chatgpt,touvron2023llama2}, more and more \textbf{LLM-based} works have been proposed  \citep{li2023angle,li2023deelm,wang2023improving} for boosting sentence embeddings significantly.

Most existing works in sentence embedding learning perform under a fixed setting, using full layers and embeddings, which limits scalability. 
To address this issue, a recent approach called Matryoshka Representation Learning (MRL) has been introduced, which allows for dynamic embedding sizes \citep{aditya2022matryoshka}.
However, while dynamic embedding size benefits downstream applications, it does not reduce computational overhead.
To overcome this limitation, we propose 2D Matryoshka Sentence Embeddings (2DMSE).
2DMSE supports elastic settings for both embedding sizes and Transformer layers, offering greater flexibility and efficiency than MRL.
It can be scaled down to smaller models with only a slight decrease in performance. 
It also effectively reduces computational overhead by choosing shallow layers. 
Its dynamic layer and embedding size make it highly versatile for various downstream applications.

\section{2D Matryoshka Sentence Embeddings Framework \matryoshka$^\mathbf{2}$}
\begin{figure*}
    \centering
    \includegraphics[width=\textwidth]{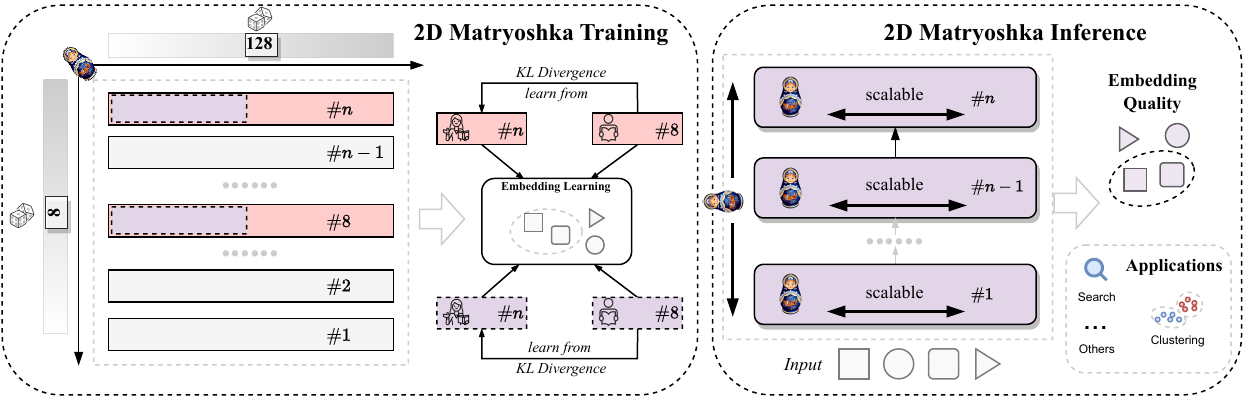}
    \caption{
    The overall framework of 2DMSE \matryoshka$^\mathbf{2}$. The left box represents the 2DMSE training stage, which involves two random processes: sampling a Transformer layer and sampling a hidden size. The selected layer and the last layer (pink rectangle) are then chosen for sentence embedding learning without scaling the hidden size. The selection of the hidden size (purple dashed rectangle) is also considered for sentence embedding learning. KL divergence is optimized during training to align the shallow layers with the last layer. The right box illustrates the inference stage, where all Transformer layers are scalable and can produce high-quality sentence embeddings for downstream applications after 2DMSE training.
    }
    \label{figure_framework}
\end{figure*}

This section elaborates on the proposed 2DMSE. The overall framework is depicted in Figure \ref{figure_framework}. We introduce the encoder backbone in Section \ref{sec::encoder} and describe scalable sentence embedding learning in Section \ref{sec::sent_learning}, followed by sentence embedding alignment in Section \ref{sec::alignment}. We present the joint learning strategy for embedding optimization in Section \ref{sec::joint_learning}.

\subsection{Encoder}
\label{sec::encoder}
We use the pretrained language model as an encoder to transform the text into dense sentence embeddings. In this work, we use BERT$_{base}$ \citep{DevlinCLT19BERT} as the backbone to encode text $x$ as follows:
\begin{equation}
    \small
    \mathbf{X}_n^d = \mathrm{BERT}_{1:n}^{cls}(x)_{1:d}  \in \mathbb{R}^d ,
\end{equation}
where $cls$ stands for the pooling strategy; we adopt the ``CLS'' embeddings as the sentence embeddings following previous works \citep{simcse_gao_2021,li2023angle}. $n\in [1, N]$ denotes the $n$-th layer of the $N$-layer Transformer backbone, and $d\in [1, D]$ represents the first $d$ dimensions in the $N$-dimensional embeddings. $n$ and $d$ largely determine the size of an encoder model, suggesting two degrees of freedom. They allow scaling the encoder model in two dimensions: the number of layers and the embedding size, which are the basis for the proposed 2DMSE.

\subsection{Scalable Sentence Embedding Learning \matryoshka}
\label{sec::sent_learning}
Following conventional approaches \cite{sbert-nils-2019,simcse_gao_2021,li2023angle}, we consistently train full-capacity embeddings from the last attention layer, $\mathbf{X}_N^D$, to ensure sentence embedding quality. The objective is as follows:
\begin{equation}
    \small
    \mathcal{L}_N^D = \mathrm{loss} (\mathbf{X}_N^D;A), \\
\end{equation}
where $\mathrm{loss}(\cdot)$ can be any loss function for sentence embedding learning, such as contrastive loss \citep{simcse_gao_2021} or AnglE loss \citep{li2023angle}. $A$ is the auxiliary information used for loss computation, such as indication for positive or negative samples or ranking information. 

Within the same training step, we randomly select a shallower Transformer layer following a uniform distribution and use its full embedding vector directly for representation learning:
\begin{equation}
    \small
    \begin{split}
        \mathcal{L}_n^D &= \mathrm{loss} (\mathbf{X}_n^D;A)\\
        n &\sim \mathcal{U}(1,N-1),
    \end{split}
\end{equation}
where $n \in [1, N)$ is the selected attention layer, and $\mathcal{U}$ denotes the uniform distribution.

To achieve scalable representation learning in 2DMSE, we apply MRL \cite{aditya2022matryoshka} to train nested low-dimensional vectors at both the last layer, $\mathbf{X}_N$:
\begin{equation}
    \small
    \begin{split}
        \mathcal{L}_N^d &= \mathrm{loss} (\mathbf{X}_N^d;A)\\
        d &\sim \mathcal{U}(1,D-1),
    \end{split}
\end{equation}
and the sampled layer, $\mathbf{X}_n$:
\begin{equation}
    \small
    \mathcal{L}_n^d = \mathrm{loss} (\mathbf{X}_n^d;A),\\
\end{equation}
where $d \in \mathbb{N}$ is the MRL embedding size and is sampled from a set of representation sizes $\mathcal{D} \subseteq [1, D-1]$. To handle various embedding dimensions efficiently, we use the geometric series with a base of $8$ and a ratio of $2$ for $\mathcal{D}$.


\subsection{Sentence Embedding Alignment \studenticon $\ \rightarrow$ \teachericon}
\label{sec::alignment}
In addition to $\mathcal{L}_N^D$, $\mathcal{L}_n^D$, $\mathcal{L}_N^d$, and $\mathcal{L}_n^d$, we adopt distribution alignment to further improve embedding performance. According to the scaling law \citep{kaplan2020scaling}, more Transformer layers have more powerful language understanding capabilities. Following this law, we align the sampled shallow layer's sentence embeddings to the last layer, thereby improving the shallow layer's performance, by minimizing their divergence:
\begin{equation}
    \small
    \mathcal{L}_{align} = \mathrm{KLDiv}(\mathcal{L}_n^D, \mathcal{L}_N^D) + \mathrm{KLDiv}(\mathcal{L}_n^d, \mathcal{L}_N^d),
\end{equation}
where $\mathrm{KLDiv}(q, p)$ denotes the Kullback-Leibler divergence, $q$ is the prediction, $p$ is the target.

\subsection{Joint Learning}
\label{sec::joint_learning}
In the end, we add up all training objectives to compose the final objective as follows:
\begin{equation}
    \small
    \mathcal{L} = \sum _L^S \lambda_L L,
\end{equation}
where $S=\{\mathcal{L}_N^D, \mathcal{L}_n^D, \mathcal{L}_N^d, \mathcal{L}_n^d,\mathcal{L}_{align}\}$ stands for the objective set. Hyperparameter $\lambda_L$ is the weight for objective $L$.

\section{Experimental Setup}
\begin{figure*}[!ht]
    \centering
    \includegraphics[width=\textwidth]{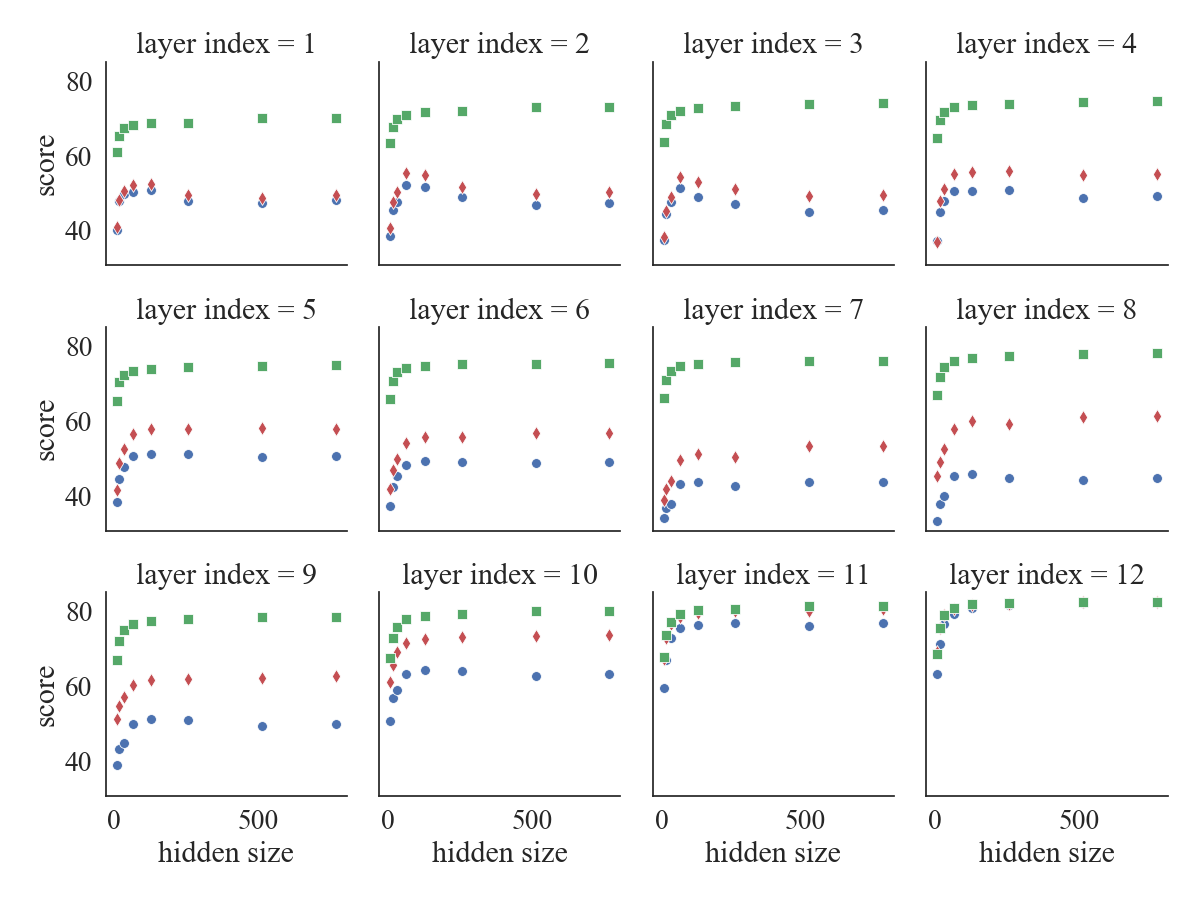}
    \caption{Results of the STS benchmark with a cascade of hidden sizes: $8 \rightarrow 16 \rightarrow 32 \rightarrow 64 \rightarrow 128 \rightarrow 256 \rightarrow 512 \rightarrow 768$ from BERT$_{base}$. The score represents the average Spearman's correlation. BERT$_{base}$ serves as the backbone for all models. The blue {$\sbullet[2]$} indicates the results of sentence embeddings from AnglE without any scalable sentence embedding learning. The red $\blacklozenge$ represents the results of matryoshka sentence embeddings. The green {$\blacksquare$} denotes the results of our proposed 2D Matryoshka Sentence Embeddings (2DMSE). The layer index $=i$ denotes the $i$-th attention layer.
    }
    \label{figure_main_results}
\end{figure*}

\paragraph{Datasets.}
We train the proposed 2DMSE on MultiNLI \citep{williams-etal-2018-broad} and SNLI \citep{bowman-etal-2015-large} datasets following previous studies and evaluate its performance on the standard STS benchmark. This benchmark comprises seven widely adopted STS datasets: STS 2012-2016 \citep{agirre-etal-2012-semeval,agirre-etal-2013-semeval,agirre-etal-2014-semeval,agirre-etal-2015-semeval,agirre-etal-2016-semeval}, SICK-R \citep{marelli-etal-2014-sick}, and STS-B \citep{cer-etal-2017-semeval}.

\paragraph{Evaluation Metrics.} 
We report Spearman's correlation coefficient, following previous studies, for a fair comparison. We compute Spearman's correlation using the SentEval toolkit \citep{conneau-kiela-2018-senteval} and present the results in the "all" setting.

\paragraph{Baselines.} 
We primarily compare the proposed 2DMSE model with the MRL model \citep{aditya2022matryoshka} to demonstrate scalability. Additionally, to showcase overall effectiveness, we compare the proposed 2DMSE model with widely adopted baselines: InferSent \citep{conneau-etal-2017-supervised}, USE \citep{cer-etal-2018-universal}, SBERT \citep{sbert-nils-2019}, SimCSE \citep{simcse_gao_2021}, and the prior STS state-of-the-art (SOTA) AnglE \citep{li2023angle}.

\paragraph{Implementation Details.}
For consistency, we utilize BERT$_{base}$ (uncased) as the backbone for all baselines. As AnglE \citep{li2023angle} has demonstrated strong performance on STS tasks, we adopt its objective as the default loss function for sentence embedding learning. The initial learning rate is set to $5e-5$, following common practices. Other hyperparameters are set following AnglE's conventions. To ensure fair comparison, we fix the random seed to $42$ for all experiments.


\section{Experimental Results}
\begin{table*}[htbp]
\small
\centering
\begin{threeparttable}
\begin{tabular}{lcccccccc}
\toprule
Model & STS12 & STS13 & STS14 & STS15 & STS16 & STS-B & SICK-R & Avg. \\
\midrule
GloVe \citep{sbert-nils-2019}  & $52.86$ & $66.75$ & $62.15$ & $72.77$ & $66.87$ & $68.03$ & $65.65$ & $65.01$ \\
USE \citep{sbert-nils-2019} & $64.49$ & $67.80$ & $64.61$ & $76.83$ & $73.18$ & $74.92$ & $76.69$ & $71.22$ \\
SBERT \citep{sbert-nils-2019} & $70.97$ & $76.53$ & $73.19$ & $79.09$ & $74.30$ & $77.03$ & $72.91$ & $74.89$ \\
SimCSE \citep{simcse_gao_2021} & $75.30$ & $84.67$ & $80.19$ & $85.40$ & $80.82$ & $84.25$ & $80.39$ & $81.57$ \\
AnglE \citep{li2023angle} & $75.09$ & $85.56$ & $80.66$ & $86.44$ & $\mathbf{82.47}$ & $85.16$ & $\mathbf{81.23}$ & $82.37$ \\
MRL ($d=768$) $\star$  & $\mathbf{75.72}$ & $\mathbf{86.79}$ & $81.89$ & $\mathbf{86.91}$ & $81.74$ & $85.50$ & $79.44$ & $82.57$ \\
\midrule
2DMSE ($n=12$, $d=768$) & $75.00$ & $86.69$ & $\mathbf{82.30}$ & $86.50$ & $82.09$ & $\mathbf{85.79}$ & $80.18$ & $\mathbf{82.65}$ \\
\bottomrule
\end{tabular}
\end{threeparttable}
\caption{
Full-capacity sentence embedding performance on the standard STS benchmark. Results $\star$ denote our implementation. BERT$_{base}$ serves as the backbone for all models.
}
\label{table_main}
\end{table*}

We discuss the main STS benchmark results in Section \ref{sec-main-result}. The ablation study investigating the significance of each component is reported in Section \ref{sec-ablation}. Furthermore, we perform an efficiency study in Section \ref{sec-efficiency} to quantify the speedup of 2DMSE in the inference stage.

\subsection{Main Results}
\label{sec-main-result}

In the main experiments, we extract the matryoshka embedding vectors of every Transformer layer from the BERT$_{base}$ backbones that are finetuned by AnglE (blue $\sbullet[2]$), AnglE with MRL (red $\blacklozenge$), and AnglE with our proposed 2DMSE (green $\blacksquare$) and test their performance on the standard STS benchmarks. For each layer, we adopt cascading vector dimensions of $\mathcal{D} = \{8, 16, 32, 64, 128, 256, 512, 768\}$. 

The dimension-wise results of all layers are visualized as dot plots in Figure \ref{figure_main_results}. Layer-wise results are presented in Figure \ref{figure_layer_scores}. Detailed results of each STS task are reported in Table \ref{table_detailed_results}, in Appendix \ref{appendix_main_results}. We also compare with strong baselines for sentence embeddings and report the results in Table \ref{table_main}.

From both Figures \ref{figure_main_results} and \ref{figure_layer_scores}, it is evident that the proposed 2DMSE can significantly improve the embedding quality of shallow Transformer layers compared to MRL. Although all the models achieve comparable results with the full-capacity embeddings from the last layer, AnglE and MRL yield inferior performance in the shallow layers and even show performance fluctuations as layers deepen. For example, AnglE achieves performance higher than $60.00$ at the eighth layer, while MRL requires ten layers to reach $60.00$. In contrast, 2DMSE achieves a score of $70.09$ in the first Transformer layer and consistently improves until the last layer with a score of $82.65$. These results indicate that 2DMSE equips each Transformer layer with promising embedding capacity, making it feasible to use the embedding vectors from shallow layers as sentence embeddings. 

Furthermore, 2DMSE extends the flexibility and multifidelity of matryoshka learning to all layers. From Figure \ref{figure_main_results}, 2DMSE produces rapid and stable performance improvements as the embedding dimension cascadingly grows. Such improvements are consistent across all layers, regardless of how the original embedding behaves, benefiting from the explicit optimization on the shallow layers. These results imply that one can utilize low-dimensional embedding vectors from intermediate layers while enjoying high embedding quality, signifying the considerable efficiency of 2DMSE in downstream tasks.

We compare 2DMSE with strong STS baselines using full-capacity sentence embedding ($n=12$, $d=768$) to test the absolute performance. Remarkably, 2DMSE outperforms these baselines. Both MRL and 2DMSE outperform the models finetuned with AnglE only, suggesting that matryoshka-style learning facilitates sentence embedding by optimizing embeddings in a nested manner.

\subsection{Ablation Study}
\label{sec-ablation}
We conduct ablation studies of the proposed 2DMSE on the standard STS benchmark. The results are presented in Table \ref{table_ablation}. From the table, we can observe that 2DMSE aided with alignment $\mathcal{L}_{align}$ and last layer learning $\mathcal{L}_N^D$ consistently outperforms other settings. This demonstrates their positive contribution to the models' performance. 
Additionally, we notice that, when the last attention layer learning $\mathcal{L}_N^D$ is omitted, performance will decrease significantly. 
This is likely because the last layer possesses strong language understanding capabilities, enhancing sentence embeddings and potentially improving the performance of sub-attention layers through alignment.

\begin{table}[]
    \centering
    \small
    \begin{tabular}{ll}
        \toprule
        Model & Avg. Spearman's Correlation \\
        \midrule
        \midrule
        \multicolumn{2}{c}{\textit{$n=12$, $d=768$}} \\
        \midrule
        2DMSE & \;\;\;\;\;\;\;\;\;$\mathbf{82.65}$ \\
        \ w/o alignment $\mathcal{L}_{align}$ & \;\;\;\;\;\;\;\;\;$82.57$ \;($-0.08$) \\
        \ w/o last layer $\mathcal{L}_N^D$ & \;\;\;\;\;\;\;\;\;$81.31$ \;($-1.34$) \\
        \midrule
        \midrule
        \multicolumn{2}{c}{\textit{$n=8$, $d=512$}} \\
        \midrule
        2DMSE &  \;\;\;\;\;\;\;\;\;$\mathbf{78.02}$ \\
        \ w/o alignment $\mathcal{L}_{align}$ & \;\;\;\;\;\;\;\;\;$77.94$ \;($-0.08$) \\
        \ w/o last layer $\mathcal{L}_N^D$ & \;\;\;\;\;\;\;\;\;$76.52$ \;($-1.50$) \\
        \midrule
        \midrule
        \multicolumn{2}{c}{\textit{$n=6$, $d=384$}} \\
        \midrule
        2DMSE & \;\;\;\;\;\;\;\;\;$\mathbf{75.21}$ \\
        \ w/o alignment $\mathcal{L}_{align}$ & \;\;\;\;\;\;\;\;\;$75.08$  \;($-0.13$) \\
        \ w/o last layer $\mathcal{L}_N^D$ & \;\;\;\;\;\;\;\;\;$74.98$ \;($-0.23$)\\
        \midrule
        \midrule
        \multicolumn{2}{c}{\textit{$n=4$, $d=256$}} \\
        \midrule
        2DMSE & \;\;\;\;\;\;\;\;\;$\mathbf{73.93}$ \\
        \ w/o alignment $\mathcal{L}_{align}$ & \;\;\;\;\;\;\;\;\;$73.69$ \;($-0.24$) \\
        \ w/o last layer $\mathcal{L}_N^D$ & \;\;\;\;\;\;\;\;\;$73.00$ \;($-0.93$)\\
        \bottomrule
    \end{tabular}
    \caption{Ablation study results of 2DMSE on the standard STS benchmark using BERT$_{base}$. $n$ denotes the number of Transformer layers, and $d$ stands for the embedding dimensions.
    }
    \label{table_ablation}
\end{table}

\subsection{Efficiency Study}
\label{sec-efficiency}
\begin{figure*}
     \centering
     \begin{subfigure}[b]{0.45\textwidth}
         \centering
         \includegraphics[width=\textwidth]{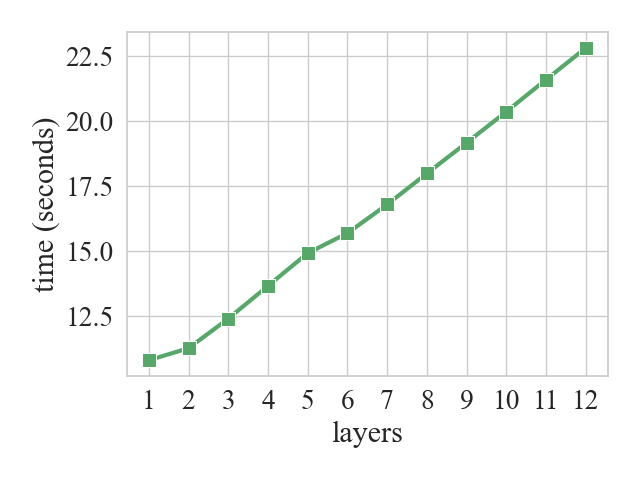}
         \caption{Inference time \textit{vs} number of layers.}
         \label{figure_time_consume}
     \end{subfigure}
     \begin{subfigure}[b]{0.45\textwidth}
         \centering
         \includegraphics[width=\textwidth]{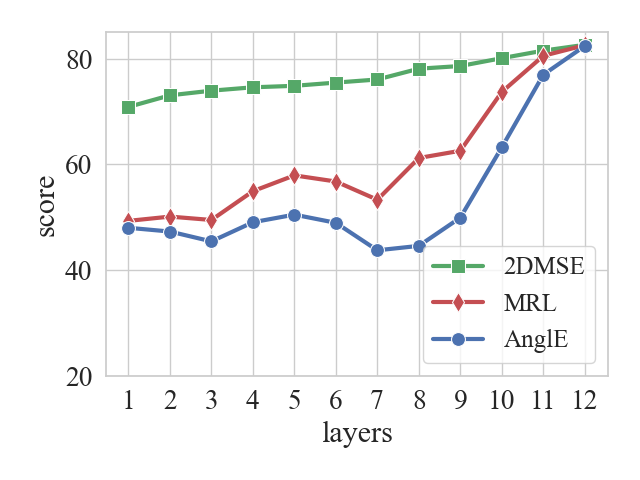}
         \caption{Score on STS \textit{vs} number of layers.}
         \label{figure_layer_scores}
     \end{subfigure}
    \caption{Subfigure (a) illustrates the time taken to use embeddings from different layers to encode the entire STS benchmarks. Subfigure (b) displays the average Spearman's correlation scores of different layers. Both (a) and (b) use an embedding size of $768$ and the standard STS benchmark dataset.}
    \label{figure_time_and_score}
\end{figure*}

To quantify the efficiency of 2DMSE during the inference stage, we record the time cost for generating embeddings at different layers for the entire STS benchmark. The results are visualized in Table \ref{figure_time_consume}. Meanwhile, we compare the performance on STS benchmarks of different learning strategies using full-capacity embeddings at each layer in Table \ref{figure_layer_scores}. The inference time linearly increases with the number of layers. For example, 2DMSE exhibits a $2.0\times$ theoretical speedup and approximately $\sim1.46\times$ real-world speedup when using the middle layer (i.e., layer $\# 6$) compared to layer $\# 12$. Regarding trade-offs in performance, 2DMSE experiences a score drop of $7.15$ using the middle layer's embedding, whereas MRL and AnglE suffer score reductions of $25.79$ and $33.44$, respectively.

\subsection{Discussion}
\label{sec-discussion}
\paragraph{Effectiveness of Two-Dimensional Matryoshka Learning.}

When comparing the sentence embedding performance on STS benchmarks using the embedding vectors from all the layers of AnglE (represented by the blue {$\sbullet[2]$} line in Figure \ref{figure_main_results}), an unexpected drop is observed as the layers deepen, specifically from layer $\#5$ to layer $\#8$. Decisive improvements are not observed until the deeper layers, from layer $\#9$ to layer $\#12$. In contrast, 2DMSE consistently yields improvements as the layers deepen. We attribute this to the rigidity of using a fixed-depth encoding pipeline, as deep learning models tend to distribute the high-level feature extraction process across all layers, even when it may not be necessary to have as many layers. Thus, matryoshka learning across layers can more effectively utilize the encoding capacity unleashed by the scaling law.


Furthermore, our experiments demonstrate that the proposed 2DMSE, applying matryoshka learning at all layers, brings further improvements over MRL, which applies matryoshka learning only at the last layer. We believe this is because 2DMSE refines all Transformer layers by interpolating coarse-to-fine-grained information across all embedding dimensions. At each layer, information tends to be concentrated at one end while maintaining a long tail at the other end. Consequently, embeddings from 2DMSE are more compact than those learned from normal fine-tuning, facilitating feature extraction in subsequent layers and ultimately improving absolute embedding performance.

\paragraph{Scalability of 2DMSE Model.}

In our main experiments, we demonstrated the superior scalability of the proposed 2DMSE compared to baselines. 
To further investigate the scaling of the 2DMSE model, we conducted an experiment where we scaled down the trained BERT$_{base}$ ($N=12$, $D=768$) 2DMSE model to BERT$_{small}$ ($N=4$, $D=512$) and BERT$_{tiny}$ ($N=12$, $D=128$) sizes. We then compared the performance of these scaled-down models with BERT$_{small}$ and BERT$_{tiny}$ models trained independently on MultiNLI + NLI. The results, presented in Table \ref{table_scaling}, show that the scaled 2DMSE consistently outperforms BERT$_{small}$ and BERT$_{tiny}$, suggesting superior scalability of the 2DMSE model.

With reduced model depth, 2DMSE provides exceptional efficiency and scalability in memory consumption and inference time, with minimal efficiency-vs-accuracy trade-off. One can simply remove the last two or three layers from the original backbone to satisfy deployment constraints.

\begin{table}[ht]
\setlength\tabcolsep{4pt}
\small
\centering
\begin{threeparttable}
\begin{tabular}{ll}
\toprule
    Model $\downarrow$  & Avg. Spearman’s Correlation  \\
    \midrule
    \midrule
    \multicolumn{2}{c}{\textit{Small Scale ($n=4$, $d=512$)}} \\
    \midrule
    BERT$_{small}$ & \;\;\;\;\;\;\;\;\;$74.01$ \\
    MRL w/ BERT$_{base}$ & \;\;\;\;\;\;\;\;\;$54.91$ \;\;\;($-19.10$) \\
    2DMSE w/ BERT$_{base}$ & \;\;\;\;\;\;\;\;\;$\mathbf{74.46}$ \;\;($+0.35$) \\
    \midrule
    \midrule
    \multicolumn{2}{c}{\textit{Tiny Scale ($n=2$, $d=128$)}} \\
    \midrule
    BERT$_{tiny}$ & \;\;\;\;\;\;\;\;\;$69.85$ \\
    MRL w/ BERT$_{base}$ & \;\;\;\;\;\;\;\;\;$54.90$ \;\;\;($-14.95$) \\
    2DMSE w/ BERT$_{base}$ & \;\;\;\;\;\;\;\;\;$\mathbf{71.64}$ \;\;($+1.79$) \\
\bottomrule
\end{tabular}
\end{threeparttable}
\caption{
Results of different model scales and their independently trained counterparts. The average Spearman's correlation of the STS Benchmark serves as the metric. $n$ denotes the number of Transformer layers, and $d$ stands for the embedding dimensions.
}
\label{table_scaling}
\end{table}

\paragraph{Discussion of Computational Overhead.}

Here, we compare the proposed 2DMSE with MRL in terms of computational overhead. MRL requires traversing all Transformer layers to produce sentence embeddings, leading to significant computational overhead. The computational complexity of MRL can be seen as $O(N)$, where $N$ is the total number of Transformer attention layers. On the other hand, 2DMSE can reduce computational overhead thanks to its scalable Transformer layer feature. Its computational complexity can be seen as $O(n)$, where $n \leq N$ represents the number of Transformer layers.

\section{Conclusion}

In this paper, we have proposed a novel sentence embedding model called 2D Matryoshka Sentence Embeddings (\matryoshka $^\mathbf{2}$) (2DMSE). 2DMSE offers enhanced scalability by accommodating encoding models of various sizes and capacities. By providing flexibility in the selection of encoding layers and their respective dimensions, our approach can adapt to different computational resources and requirements. This scalability empowers researchers and practitioners to efficiently leverage 2DMSE in diverse settings.

Extensive experiments on STS benchmarks have demonstrated that 2DMSE consistently outperforms baselines and exhibits superior scalability, making our approach well-suited for a wide range of downstream applications.




\bibliography{anthology,custom}

\appendix
\section{Main Results}
\label{appendix_main_results}
The detailed results of Figure \ref{figure_main_results} are presented in Table \ref{table_detailed_results}.

\onecolumn
\begin{longtable}{lcccccccc}
\toprule
\textbf{Model} & \textbf{STS12} & \textbf{STS13} & \textbf{STS14} & \textbf{STS15} & \textbf{STS16} & \textbf{STS-B} & \textbf{Sick-R} & \textbf{Avg.} \\
\midrule
\midrule
\multicolumn{9}{c}{$\#$ Layer $n=1$}\\
\midrule
AnglE ($d=8$) & $32.24$ & $39.92$ & $37.80$ & $39.93$ & $45.66$ & $38.48$ & $45.86$ & $39.98$\\
MRL ($d=8$) & $32.21$ & $42.39$ & $39.60$ & $40.18$ & $44.65$ & $40.05$ & $47.25$ & $40.90$\\
2DMSE ($d=8$) & $58.34$ & $60.35$ & $56.99$ & $65.13$ & $58.81$ & $61.66$ & $64.97$ & $60.89$\\
\midrule
AnglE ($d=16$) & $44.06$ & $47.76$ & $43.39$ & $49.39$ & $53.41$ & $44.34$ & $52.65$ & $47.86$\\
MRL ($d=16$) & $39.93$ & $49.95$ & $45.93$ & $47.35$ & $54.78$ & $46.25$ & $52.80$ & $48.14$\\
2DMSE ($d=16$) & $61.32$ & $66.25$ & $61.35$ & $70.14$ & $63.58$ & $66.90$ & $67.97$ & $65.36$\\
\midrule
AnglE ($d=32$) & $48.26$ & $49.72$ & $44.03$ & $52.76$ & $53.97$ & $46.48$ & $52.92$ & $49.73$\\
MRL ($d=32$) & $47.33$ & $51.83$ & $46.37$ & $51.37$ & $56.37$ & $47.45$ & $53.15$ & $50.55$\\
2DMSE ($d=32$) & $62.39$ & $68.29$ & $62.89$ & $72.88$ & $67.24$ & $69.17$ & $67.96$ & $67.26$\\
\midrule
AnglE ($d=64$) & $49.24$ & $47.23$ & $42.65$ & $54.94$ & $56.40$ & $48.35$ & $53.86$ & $50.38$\\
MRL ($d=64$) & $50.49$ & $49.85$ & $45.69$ & $55.95$ & $58.35$ & $50.57$ & $54.16$ & $52.15$\\
2DMSE ($d=64$) & $63.78$ & $68.48$ & $63.15$ & $74.56$ & $68.35$ & $70.35$ & $69.04$ & $68.24$\\
\midrule
AnglE ($d=128$) & $50.27$ & $48.38$ & $44.29$ & $53.91$ & $56.84$ & $47.92$ & $53.09$ & $50.67$\\
MRL ($d=128$) & $51.43$ & $51.08$ & $47.15$ & $54.72$ & $58.95$ & $49.86$ & $53.55$ & $52.39$\\
2DMSE ($d=128$) & $64.32$ & $69.36$ & $63.61$ & $74.80$ & $69.39$ & $70.74$ & $69.31$ & $68.79$\\
\midrule
AnglE ($d=256$) & $46.75$ & $47.24$ & $43.13$ & $50.02$ & $55.00$ & $43.92$ & $49.36$ & $47.92$\\
MRL ($d=256$) & $47.55$ & $49.31$ & $45.33$ & $50.81$ & $56.91$ & $45.73$ & $49.81$ & $49.35$\\
2DMSE ($d=256$) & $64.10$ & $69.90$ & $63.27$ & $75.04$ & $69.88$ & $70.20$ & $68.91$ & $68.76$\\
\midrule
AnglE ($d=512$) & $46.33$ & $44.56$ & $40.17$ & $51.29$ & $52.66$ & $44.03$ & $51.95$ & $47.28$\\
MRL ($d=512$) & $47.95$ & $45.33$ & $41.89$ & $52.60$ & $53.50$ & $46.03$ & $53.07$ & $48.62$\\
2DMSE ($d=512$) & $65.02$ & $70.75$ & $65.41$ & $77.85$ & $70.33$ & $71.30$ & $69.81$ & $70.07$\\
\midrule
AnglE ($d=768$) & $47.15$ & $45.33$ & $41.25$ & $52.06$ & $53.13$ & $44.79$ & $52.41$ & $48.02$\\
MRL ($d=768$) & $48.66$ & $46.16$ & $42.93$ & $53.37$ & $53.98$ & $46.65$ & $53.55$ & $49.33$\\
2DMSE ($d=768$) & $64.89$ & $70.78$ & $65.34$ & $77.74$ & $70.54$ & $71.30$ & $70.04$ & $70.09$\\
\midrule
\midrule
\multicolumn{9}{c}{$\#$ Layer $n=2$}\\
\midrule
AnglE ($d=8$) & $26.41$ & $37.68$ & $34.70$ & $41.84$ & $46.83$ & $36.67$ & $45.03$ & $38.45$\\
MRL ($d=8$) & $29.58$ & $42.70$ & $37.05$ & $43.20$ & $45.80$ & $39.24$ & $46.68$ & $40.61$\\
2DMSE ($d=8$) & $60.73$ & $62.66$ & $59.75$ & $68.77$ & $60.48$ & $64.29$ & $66.43$ & $63.30$\\
\midrule
AnglE ($d=16$) & $35.22$ & $45.28$ & $41.20$ & $49.50$ & $52.42$ & $42.78$ & $51.55$ & $45.42$\\
MRL ($d=16$) & $37.61$ & $49.53$ & $44.17$ & $51.17$ & $53.12$ & $46.03$ & $52.24$ & $47.70$\\
2DMSE ($d=16$) & $63.77$ & $67.93$ & $63.96$ & $73.50$ & $65.34$ & $69.01$ & $69.64$ & $67.59$\\
\midrule
AnglE ($d=32$) & $42.39$ & $46.89$ & $43.10$ & $49.60$ & $54.16$ & $45.07$ & $52.18$ & $47.63$\\
MRL ($d=32$) & $46.31$ & $51.11$ & $45.63$ & $51.78$ & $55.71$ & $47.44$ & $53.36$ & $50.19$\\
2DMSE ($d=32$) & $65.80$ & $70.98$ & $65.93$ & $75.73$ & $69.30$ & $71.05$ & $70.60$ & $69.91$\\
\midrule
AnglE ($d=64$) & $46.12$ & $51.18$ & $45.17$ & $58.10$ & $59.10$ & $51.09$ & $54.82$ & $52.23$\\
MRL ($d=64$) & $51.33$ & $54.94$ & $49.07$ & $60.30$ & $60.64$ & $54.90$ & $56.36$ & $55.36$\\
2DMSE ($d=64$) & $66.19$ & $72.15$ & $66.83$ & $76.90$ & $70.49$ & $71.94$ & $72.07$ & $70.94$\\
\midrule
AnglE ($d=128$) & $46.03$ & $50.75$ & $45.26$ & $57.05$ & $59.80$ & $48.54$ & $54.45$ & $51.70$\\
MRL ($d=128$) & $50.30$ & $54.40$ & $49.37$ & $59.59$ & $61.32$ & $52.95$ & $56.39$ & $54.90$\\
2DMSE ($d=128$) & $66.46$ & $72.77$ & $67.40$ & $77.54$ & $71.85$ & $72.65$ & $72.84$ & $71.64$\\
\midrule
AnglE ($d=256$) & $44.39$ & $49.07$ & $43.65$ & $52.23$ & $57.39$ & $44.65$ & $51.49$ & $48.98$\\
MRL ($d=256$) & $47.33$ & $51.88$ & $46.74$ & $54.53$ & $58.93$ & $48.03$ & $53.00$ & $51.49$\\
2DMSE ($d=256$) & $66.15$ & $73.24$ & $67.36$ & $77.85$ & $72.71$ & $73.00$ & $73.18$ & $71.93$\\
\midrule
AnglE ($d=512$) & $44.33$ & $45.13$ & $39.26$ & $51.23$ & $52.84$ & $42.96$ & $52.04$ & $46.83$\\
MRL ($d=512$) & $48.28$ & $47.80$ & $42.80$ & $54.25$ & $54.48$ & $46.54$ & $53.93$ & $49.73$\\
2DMSE ($d=512$) & $67.76$ & $73.08$ & $69.30$ & $80.03$ & $73.38$ & $74.25$ & $73.37$ & $73.02$\\
\midrule
AnglE ($d=768$) & $44.79$ & $45.59$ & $39.98$ & $51.54$ & $53.18$ & $43.67$ & $52.39$ & $47.31$\\
MRL ($d=768$) & $48.62$ & $48.13$ & $43.47$ & $54.53$ & $54.83$ & $47.12$ & $54.29$ & $50.14$\\
2DMSE ($d=768$) & $67.68$ & $73.53$ & $69.17$ & $79.80$ & $73.70$ & $74.21$ & $73.52$ & $73.09$\\
\midrule
\midrule
\multicolumn{9}{c}{$\#$ Layer $n=3$}\\
\midrule
AnglE ($d=8$) & $25.87$ & $35.33$ & $34.69$ & $40.32$ & $45.56$ & $35.79$ & $43.80$ & $37.34$\\
MRL ($d=8$) & $23.52$ & $39.87$ & $36.65$ & $42.11$ & $44.18$ & $34.68$ & $45.98$ & $38.14$\\
2DMSE ($d=8$) & $60.73$ & $61.75$ & $61.37$ & $67.90$ & $62.11$ & $65.01$ & $67.34$ & $63.74$\\
\midrule
AnglE ($d=16$) & $32.79$ & $43.00$ & $40.69$ & $50.57$ & $51.79$ & $42.05$ & $49.93$ & $44.40$\\
MRL ($d=16$) & $31.63$ & $46.16$ & $42.50$ & $50.85$ & $51.00$ & $42.01$ & $51.92$ & $45.15$\\
2DMSE ($d=16$) & $63.46$ & $68.55$ & $65.52$ & $73.82$ & $67.16$ & $69.95$ & $70.65$ & $68.44$\\
\midrule
AnglE ($d=32$) & $39.20$ & $45.35$ & $42.12$ & $53.29$ & $54.78$ & $44.62$ & $52.88$ & $47.46$\\
MRL ($d=32$) & $39.73$ & $48.94$ & $43.60$ & $54.28$ & $55.41$ & $45.77$ & $54.61$ & $48.91$\\
2DMSE ($d=32$) & $65.74$ & $72.08$ & $67.72$ & $76.28$ & $70.60$ & $72.06$ & $71.79$ & $70.90$\\
\midrule
AnglE ($d=64$) & $45.79$ & $49.75$ & $43.31$ & $57.32$ & $58.66$ & $49.40$ & $54.26$ & $51.21$\\
MRL ($d=64$) & $49.82$ & $52.86$ & $46.75$ & $60.49$ & $60.53$ & $53.24$ & $56.47$ & $54.31$\\
2DMSE ($d=64$) & $66.45$ & $73.60$ & $68.75$ & $77.53$ & $71.74$ & $73.08$ & $72.92$ & $72.01$\\
\midrule
AnglE ($d=128$) & $43.10$ & $47.13$ & $41.10$ & $54.51$ & $58.57$ & $45.55$ & $51.79$ & $48.82$\\
MRL ($d=128$) & $47.55$ & $51.36$ & $45.65$ & $59.33$ & $61.22$ & $51.13$ & $55.06$ & $53.04$\\
2DMSE ($d=128$) & $67.31$ & $74.31$ & $69.21$ & $78.07$ & $72.75$ & $73.82$ & $73.37$ & $72.69$\\
\midrule
AnglE ($d=256$) & $43.19$ & $46.29$ & $40.87$ & $50.70$ & $56.22$ & $42.34$ & $49.71$ & $47.05$\\
MRL ($d=256$) & $46.75$ & $50.57$ & $45.16$ & $55.27$ & $59.16$ & $48.11$ & $53.15$ & $51.17$\\
2DMSE ($d=256$) & $67.60$ & $74.81$ & $69.30$ & $78.63$ & $73.67$ & $74.65$ & $73.86$ & $73.22$\\
\midrule
AnglE ($d=512$) & $42.00$ & $42.25$ & $37.23$ & $49.34$ & $52.34$ & $40.89$ & $50.07$ & $44.87$\\
MRL ($d=512$) & $45.88$ & $47.14$ & $42.03$ & $54.43$ & $54.82$ & $45.88$ & $53.39$ & $49.08$\\
2DMSE ($d=512$) & $68.35$ & $74.38$ & $70.41$ & $80.21$ & $74.28$ & $75.73$ & $73.72$ & $73.87$\\
\midrule
AnglE ($d=768$) & $42.88$ & $42.78$ & $37.95$ & $49.71$ & $52.73$ & $41.78$ & $50.33$ & $45.45$\\
MRL ($d=768$) & $46.67$ & $47.28$ & $42.54$ & $54.72$ & $55.16$ & $46.57$ & $53.55$ & $49.50$\\
2DMSE ($d=768$) & $68.48$ & $74.73$ & $70.21$ & $80.07$ & $74.49$ & $75.73$ & $74.08$ & $73.97$\\
\midrule
\midrule
\multicolumn{9}{c}{$\#$ Layer $n=4$}\\
\midrule
AnglE ($d=8$) & $30.39$ & $37.34$ & $32.51$ & $40.76$ & $46.53$ & $33.32$ & $39.21$ & $37.15$\\
MRL ($d=8$) & $21.68$ & $39.76$ & $34.19$ & $43.94$ & $47.53$ & $31.78$ & $40.00$ & $36.98$\\
2DMSE ($d=8$) & $61.90$ & $63.42$ & $62.02$ & $68.96$ & $62.76$ & $65.77$ & $67.96$ & $64.68$\\
\midrule
AnglE ($d=16$) & $34.34$ & $46.17$ & $40.84$ & $51.61$ & $51.47$ & $43.94$ & $46.23$ & $44.94$\\
MRL ($d=16$) & $35.84$ & $49.26$ & $45.02$ & $55.65$ & $56.23$ & $46.03$ & $47.12$ & $47.88$\\
2DMSE ($d=16$) & $64.89$ & $69.83$ & $66.36$ & $74.53$ & $68.52$ & $71.18$ & $71.09$ & $69.49$\\
\midrule
AnglE ($d=32$) & $37.42$ & $48.24$ & $41.58$ & $55.34$ & $55.62$ & $46.00$ & $50.45$ & $47.81$\\
MRL ($d=32$) & $40.99$ & $51.02$ & $44.88$ & $58.37$ & $60.02$ & $49.49$ & $51.84$ & $50.94$\\
2DMSE ($d=32$) & $66.73$ & $73.48$ & $68.47$ & $76.19$ & $71.65$ & $73.30$ & $72.58$ & $71.77$\\
\midrule
AnglE ($d=64$) & $44.42$ & $50.62$ & $42.45$ & $57.06$ & $57.78$ & $49.80$ & $52.27$ & $50.63$\\
MRL ($d=64$) & $50.17$ & $53.86$ & $47.07$ & $62.08$ & $61.08$ & $55.43$ & $55.48$ & $55.02$\\
2DMSE ($d=64$) & $67.72$ & $74.83$ & $69.54$ & $77.94$ & $72.81$ & $74.11$ & $73.57$ & $72.93$\\
\midrule
AnglE ($d=128$) & $44.28$ & $48.90$ & $41.88$ & $57.51$ & $59.38$ & $49.58$ & $52.32$ & $50.55$\\
MRL ($d=128$) & $50.44$ & $53.38$ & $47.18$ & $63.38$ & $63.37$ & $55.93$ & $56.38$ & $55.72$\\
2DMSE ($d=128$) & $68.47$ & $75.05$ & $70.04$ & $78.41$ & $73.39$ & $74.70$ & $74.22$ & $73.47$\\
\midrule
AnglE ($d=256$) & $45.07$ & $49.23$ & $42.48$ & $57.45$ & $59.84$ & $48.79$ & $51.91$ & $50.68$\\
MRL ($d=256$) & $50.44$ & $53.89$ & $48.03$ & $63.01$ & $64.09$ & $55.32$ & $56.11$ & $55.84$\\
2DMSE ($d=256$) & $68.95$ & $75.45$ & $69.98$ & $79.01$ & $74.19$ & $75.40$ & $74.54$ & $73.93$\\
\midrule
AnglE ($d=512$) & $44.78$ & $45.94$ & $40.20$ & $54.44$ & $56.76$ & $47.07$ & $51.65$ & $48.69$\\
MRL ($d=512$) & $51.34$ & $52.35$ & $46.72$ & $61.16$ & $61.15$ & $55.08$ & $56.56$ & $54.91$\\
2DMSE ($d=512$) & $69.44$ & $75.25$ & $70.97$ & $80.29$ & $74.62$ & $76.40$ & $74.24$ & $74.46$\\
\midrule
AnglE ($d=768$) & $45.38$ & $46.24$ & $40.73$ & $54.84$ & $57.07$ & $47.49$ & $51.71$ & $49.07$\\
MRL ($d=768$) & $51.68$ & $52.20$ & $46.90$ & $61.24$ & $61.16$ & $55.01$ & $56.44$ & $54.95$\\
2DMSE ($d=768$) & $69.77$ & $75.56$ & $70.80$ & $80.23$ & $74.89$ & $76.45$ & $74.56$ & $74.61$\\
\midrule
\midrule
\multicolumn{9}{c}{$\#$ Layer $n=5$}\\
\midrule
AnglE ($d=8$) & $29.46$ & $40.52$ & $35.44$ & $43.85$ & $46.16$ & $30.69$ & $41.24$ & $38.19$\\
MRL ($d=8$) & $25.81$ & $42.71$ & $36.61$ & $50.08$ & $50.04$ & $36.01$ & $48.81$ & $41.44$\\
2DMSE ($d=8$) & $64.49$ & $63.63$ & $62.30$ & $69.04$ & $63.76$ & $66.43$ & $68.34$ & $65.43$\\
\midrule
AnglE ($d=16$) & $38.29$ & $47.23$ & $39.62$ & $52.33$ & $48.34$ & $37.06$ & $47.73$ & $44.37$\\
MRL ($d=16$) & $38.18$ & $49.70$ & $43.84$ & $58.91$ & $55.19$ & $43.45$ & $51.77$ & $48.72$\\
2DMSE ($d=16$) & $67.51$ & $70.51$ & $67.08$ & $74.60$ & $69.42$ & $71.79$ & $71.09$ & $70.29$\\
\midrule
AnglE ($d=32$) & $38.13$ & $48.52$ & $40.71$ & $56.08$ & $54.90$ & $41.73$ & $53.54$ & $47.66$\\
MRL ($d=32$) & $41.40$ & $52.11$ & $45.72$ & $61.31$ & $60.16$ & $49.23$ & $57.34$ & $52.47$\\
2DMSE ($d=32$) & $68.16$ & $73.89$ & $69.29$ & $75.93$ & $72.00$ & $73.54$ & $72.44$ & $72.18$\\
\midrule
AnglE ($d=64$) & $43.72$ & $50.22$ & $42.17$ & $58.35$ & $57.21$ & $47.65$ & $55.44$ & $50.68$\\
MRL ($d=64$) & $49.75$ & $55.55$ & $48.85$ & $64.67$ & $61.99$ & $56.96$ & $58.54$ & $56.62$\\
2DMSE ($d=64$) & $68.98$ & $74.99$ & $70.10$ & $77.61$ & $73.36$ & $74.35$ & $73.69$ & $73.30$\\
\midrule
AnglE ($d=128$) & $42.82$ & $50.30$ & $42.01$ & $58.98$ & $59.50$ & $48.06$ & $55.64$ & $51.04$\\
MRL ($d=128$) & $50.39$ & $56.18$ & $49.34$ & $65.65$ & $64.30$ & $58.77$ & $59.73$ & $57.77$\\
2DMSE ($d=128$) & $69.24$ & $75.59$ & $70.70$ & $78.02$ & $74.10$ & $75.05$ & $74.29$ & $73.86$\\
\midrule
AnglE ($d=256$) & $41.87$ & $52.18$ & $42.69$ & $59.01$ & $60.42$ & $46.93$ & $54.83$ & $51.13$\\
MRL ($d=256$) & $49.30$ & $57.49$ & $49.84$ & $65.60$ & $65.60$ & $57.55$ & $59.22$ & $57.80$\\
2DMSE ($d=256$) & $69.49$ & $76.07$ & $70.76$ & $78.96$ & $74.87$ & $75.95$ & $74.48$ & $74.37$\\
\midrule
AnglE ($d=512$) & $44.88$ & $49.86$ & $41.57$ & $57.19$ & $58.65$ & $45.63$ & $54.58$ & $50.34$\\
MRL ($d=512$) & $53.46$ & $56.87$ & $49.66$ & $64.76$ & $64.13$ & $57.73$ & $59.50$ & $58.02$\\
2DMSE ($d=512$) & $69.76$ & $75.66$ & $71.64$ & $80.03$ & $75.14$ & $76.61$ & $73.98$ & $74.69$\\
\midrule
AnglE ($d=768$) & $44.84$ & $50.22$ & $41.77$ & $57.54$ & $58.84$ & $45.74$ & $54.64$ & $50.51$\\
MRL ($d=768$) & $53.08$ & $56.95$ & $49.61$ & $64.85$ & $64.22$ & $57.40$ & $59.49$ & $57.94$\\
2DMSE ($d=768$) & $70.12$ & $76.01$ & $71.45$ & $80.08$ & $75.43$ & $76.75$ & $74.40$ & $74.89$\\
\midrule
\midrule
\multicolumn{9}{c}{$\#$ Layer $n=6$}\\
\midrule
AnglE ($d=8$) & $28.48$ & $42.25$ & $34.83$ & $40.93$ & $43.27$ & $29.63$ & $41.93$ & $37.33$\\
MRL ($d=8$) & $28.02$ & $40.18$ & $36.33$ & $51.92$ & $49.60$ & $36.53$ & $50.40$ & $41.85$\\
2DMSE ($d=8$) & $64.65$ & $64.15$ & $63.02$ & $69.65$ & $63.99$ & $67.01$ & $68.34$ & $65.83$\\
\midrule
AnglE ($d=16$) & $35.74$ & $46.28$ & $38.05$ & $50.36$ & $45.00$ & $34.08$ & $47.19$ & $42.39$\\
MRL ($d=16$) & $36.36$ & $47.59$ & $41.05$ & $57.14$ & $52.20$ & $40.97$ & $52.79$ & $46.87$\\
2DMSE ($d=16$) & $67.80$ & $71.56$ & $67.91$ & $74.52$ & $69.61$ & $72.79$ & $71.47$ & $70.81$\\
\midrule
AnglE ($d=32$) & $34.78$ & $48.15$ & $38.93$ & $52.59$ & $52.01$ & $38.14$ & $52.57$ & $45.31$\\
MRL ($d=32$) & $38.46$ & $50.02$ & $42.32$ & $58.25$ & $57.30$ & $45.18$ & $57.08$ & $49.80$\\
2DMSE ($d=32$) & $68.69$ & $75.42$ & $70.47$ & $76.17$ & $72.74$ & $74.85$ & $72.92$ & $73.04$\\
\midrule
AnglE ($d=64$) & $40.02$ & $49.26$ & $39.91$ & $55.29$ & $55.20$ & $44.21$ & $53.73$ & $48.23$\\
MRL ($d=64$) & $46.77$ & $53.69$ & $46.23$ & $61.57$ & $60.01$ & $53.29$ & $57.38$ & $54.13$\\
2DMSE ($d=64$) & $69.52$ & $76.74$ & $71.25$ & $77.92$ & $74.26$ & $76.01$ & $74.18$ & $74.27$\\
\midrule
AnglE ($d=128$) & $39.84$ & $50.11$ & $40.31$ & $56.86$ & $57.52$ & $45.48$ & $54.88$ & $49.29$\\
MRL ($d=128$) & $47.09$ & $55.48$ & $47.40$ & $62.55$ & $62.70$ & $55.97$ & $59.22$ & $55.77$\\
2DMSE ($d=128$) & $69.70$ & $77.20$ & $71.92$ & $78.39$ & $74.77$ & $76.59$ & $74.25$ & $74.69$\\
\midrule
AnglE ($d=256$) & $38.08$ & $51.23$ & $40.61$ & $56.61$ & $58.39$ & $43.86$ & $53.71$ & $48.93$\\
MRL ($d=256$) & $44.91$ & $56.81$ & $47.94$ & $62.63$ & $64.11$ & $54.68$ & $59.13$ & $55.74$\\
2DMSE ($d=256$) & $70.00$ & $77.72$ & $72.20$ & $79.38$ & $75.28$ & $77.38$ & $74.09$ & $75.15$\\
\midrule
AnglE ($d=512$) & $42.91$ & $48.50$ & $39.62$ & $56.12$ & $57.36$ & $43.50$ & $53.14$ & $48.74$\\
MRL ($d=512$) & $50.92$ & $56.20$ & $48.09$ & $63.95$ & $63.31$ & $55.88$ & $59.00$ & $56.76$\\
2DMSE ($d=512$) & $70.75$ & $76.91$ & $72.84$ & $80.40$ & $75.32$ & $77.84$ & $73.36$ & $75.35$\\
\midrule
AnglE ($d=768$) & $42.84$ & $48.97$ & $39.89$ & $56.34$ & $57.47$ & $43.81$ & $53.40$ & $48.96$\\
MRL ($d=768$) & $50.59$ & $56.46$ & $48.19$ & $63.74$ & $63.49$ & $55.85$ & $59.15$ & $56.78$\\
2DMSE ($d=768$) & $70.87$ & $77.26$ & $72.69$ & $80.41$ & $75.56$ & $78.04$ & $73.65$ & $75.50$\\
\midrule
\midrule
\multicolumn{9}{c}{$\#$ Layer $n=7$}\\
\midrule
AnglE ($d=8$) & $17.23$ & $40.98$ & $32.35$ & $39.12$ & $41.39$ & $26.25$ & $40.54$ & $33.98$\\
MRL ($d=8$) & $23.71$ & $41.64$ & $34.63$ & $45.35$ & $47.47$ & $32.45$ & $45.74$ & $38.71$\\
2DMSE ($d=8$) & $64.42$ & $63.40$ & $63.22$ & $70.03$ & $64.34$ & $67.94$ & $68.82$ & $66.02$\\
\midrule
AnglE ($d=16$) & $21.76$ & $42.24$ & $33.69$ & $43.24$ & $45.65$ & $28.69$ & $41.71$ & $36.71$\\
MRL ($d=16$) & $27.36$ & $45.74$ & $37.19$ & $48.37$ & $51.68$ & $34.14$ & $47.43$ & $41.70$\\
2DMSE ($d=16$) & $67.65$ & $70.93$ & $68.19$ & $74.17$ & $69.94$ & $74.05$ & $72.41$ & $71.05$\\
\midrule
AnglE ($d=32$) & $16.48$ & $46.16$ & $32.34$ & $41.39$ & $52.90$ & $28.63$ & $46.37$ & $37.75$\\
MRL ($d=32$) & $27.94$ & $48.58$ & $38.10$ & $48.56$ & $56.44$ & $36.58$ & $50.84$ & $43.86$\\
2DMSE ($d=32$) & $68.78$ & $75.10$ & $70.96$ & $76.22$ & $72.59$ & $75.98$ & $73.99$ & $73.37$\\
\midrule
AnglE ($d=64$) & $30.52$ & $47.70$ & $35.52$ & $48.37$ & $53.99$ & $37.54$ & $48.36$ & $43.14$\\
MRL ($d=64$) & $38.91$ & $52.15$ & $42.77$ & $54.97$ & $58.84$ & $46.25$ & $53.19$ & $49.58$\\
2DMSE ($d=64$) & $70.04$ & $76.63$ & $71.82$ & $78.05$ & $74.25$ & $76.85$ & $75.16$ & $74.69$\\
\midrule
AnglE ($d=128$) & $30.77$ & $47.75$ & $35.43$ & $49.94$ & $54.96$ & $37.41$ & $48.81$ & $43.58$\\
MRL ($d=128$) & $40.50$ & $54.16$ & $44.01$ & $56.51$ & $60.75$ & $48.22$ & $53.77$ & $51.13$\\
2DMSE ($d=128$) & $70.47$ & $77.34$ & $72.67$ & $78.83$ & $74.90$ & $77.64$ & $75.42$ & $75.32$\\
\midrule
AnglE ($d=256$) & $27.84$ & $48.57$ & $35.28$ & $48.57$ & $55.05$ & $35.41$ & $48.19$ & $42.70$\\
MRL ($d=256$) & $36.14$ & $55.30$ & $43.89$ & $55.23$ & $61.62$ & $45.86$ & $53.68$ & $50.25$\\
2DMSE ($d=256$) & $70.62$ & $77.73$ & $72.74$ & $79.67$ & $75.35$ & $78.19$ & $75.33$ & $75.66$\\
\midrule
AnglE ($d=512$) & $35.35$ & $45.40$ & $35.31$ & $50.22$ & $53.78$ & $36.90$ & $48.42$ & $43.63$\\
MRL ($d=512$) & $46.77$ & $53.88$ & $46.21$ & $59.74$ & $61.26$ & $51.02$ & $54.45$ & $53.33$\\
2DMSE ($d=512$) & $71.66$ & $77.30$ & $73.37$ & $81.01$ & $75.47$ & $78.57$ & $74.84$ & $76.03$\\
\midrule
AnglE ($d=768$) & $35.08$ & $45.78$ & $35.57$ & $50.49$ & $53.83$ & $36.96$ & $48.59$ & $43.76$\\
MRL ($d=768$) & $45.87$ & $54.56$ & $46.34$ & $59.46$ & $61.50$ & $50.96$ & $54.61$ & $53.33$\\
2DMSE ($d=768$) & $71.57$ & $77.64$ & $73.29$ & $80.92$ & $75.63$ & $78.65$ & $74.93$ & $76.09$\\
\midrule
\midrule
\multicolumn{9}{c}{$\#$ Layer $n=8$}\\
\midrule
AnglE ($d=8$) & $11.53$ & $38.60$ & $32.96$ & $41.79$ & $38.23$ & $24.76$ & $43.95$ & $33.12$\\
MRL ($d=8$) & $37.35$ & $46.52$ & $39.84$ & $46.34$ & $51.98$ & $43.43$ & $51.31$ & $45.25$\\
2DMSE ($d=8$) & $64.24$ & $65.14$ & $63.67$ & $70.42$ & $66.21$ & $68.87$ & $69.04$ & $66.80$\\
\midrule
AnglE ($d=16$) & $21.80$ & $41.79$ & $33.74$ & $44.52$ & $45.35$ & $30.67$ & $45.88$ & $37.68$\\
MRL ($d=16$) & $40.91$ & $51.06$ & $42.62$ & $50.55$ & $57.65$ & $46.04$ & $53.56$ & $48.91$\\
2DMSE ($d=16$) & $67.52$ & $71.78$ & $68.82$ & $74.68$ & $71.27$ & $74.52$ & $73.13$ & $71.67$\\
\midrule
AnglE ($d=32$) & $21.15$ & $46.99$ & $34.47$ & $42.67$ & $52.15$ & $32.00$ & $49.24$ & $39.81$\\
MRL ($d=32$) & $44.44$ & $55.63$ & $45.75$ & $52.72$ & $62.59$ & $50.08$ & $56.88$ & $52.58$\\
2DMSE ($d=32$) & $69.19$ & $76.04$ & $71.99$ & $76.98$ & $74.27$ & $76.98$ & $74.94$ & $74.34$\\
\midrule
AnglE ($d=64$) & $33.96$ & $49.19$ & $37.35$ & $49.08$ & $55.31$ & $40.49$ & $51.20$ & $45.23$\\
MRL ($d=64$) & $53.37$ & $58.65$ & $49.70$ & $59.30$ & $65.70$ & $58.12$ & $59.77$ & $57.80$\\
2DMSE ($d=64$) & $70.92$ & $77.86$ & $73.25$ & $79.10$ & $76.19$ & $78.38$ & $76.12$ & $75.97$\\
\midrule
AnglE ($d=128$) & $32.35$ & $50.38$ & $37.86$ & $51.07$ & $56.60$ & $40.41$ & $51.67$ & $45.76$\\
MRL ($d=128$) & $55.46$ & $60.59$ & $51.38$ & $62.30$ & $67.98$ & $61.03$ & $61.11$ & $59.98$\\
2DMSE ($d=128$) & $71.50$ & $78.91$ & $74.31$ & $80.46$ & $76.84$ & $79.70$ & $76.71$ & $76.92$\\
\midrule
AnglE ($d=256$) & $29.43$ & $50.44$ & $37.43$ & $49.84$ & $56.44$ & $37.95$ & $51.12$ & $44.66$\\
MRL ($d=256$) & $52.09$ & $60.77$ & $51.49$ & $61.04$ & $68.14$ & $60.07$ & $61.15$ & $59.25$\\
2DMSE ($d=256$) & $71.73$ & $79.54$ & $74.73$ & $81.28$ & $77.46$ & $80.55$ & $76.91$ & $77.46$\\
\midrule
AnglE ($d=512$) & $35.70$ & $46.06$ & $35.63$ & $50.10$ & $54.54$ & $37.94$ & $49.47$ & $44.21$\\
MRL ($d=512$) & $58.02$ & $59.07$ & $52.68$ & $65.13$ & $67.73$ & $63.13$ & $60.73$ & $60.93$\\
2DMSE ($d=512$) & $72.95$ & $79.05$ & $75.94$ & $82.64$ & $77.77$ & $81.43$ & $76.43$ & $78.03$\\
\midrule
AnglE ($d=768$) & $35.43$ & $46.69$ & $36.10$ & $50.80$ & $54.96$ & $38.20$ & $50.03$ & $44.60$\\
MRL ($d=768$) & $57.56$ & $59.83$ & $53.01$ & $65.31$ & $68.12$ & $63.57$ & $61.19$ & $61.23$\\
2DMSE ($d=768$) & $72.93$ & $79.57$ & $75.93$ & $82.52$ & $77.92$ & $81.47$ & $76.60$ & $78.13$\\
\midrule
\midrule
\multicolumn{9}{c}{$\#$ Layer $n=9$}\\
\midrule
AnglE ($d=8$) & $29.60$ & $36.46$ & $35.65$ & $50.23$ & $40.42$ & $32.26$ & $47.43$ & $38.86$\\
MRL ($d=8$) & $47.80$ & $51.30$ & $47.09$ & $52.16$ & $55.71$ & $50.79$ & $54.38$ & $51.32$\\
2DMSE ($d=8$) & $64.21$ & $65.65$ & $64.37$ & $71.21$ & $66.51$ & $69.43$ & $68.43$ & $67.12$\\
\midrule
AnglE ($d=16$) & $37.21$ & $39.61$ & $37.10$ & $51.63$ & $48.19$ & $37.87$ & $51.21$ & $43.26$\\
MRL ($d=16$) & $50.95$ & $55.35$ & $49.99$ & $55.93$ & $60.89$ & $52.87$ & $56.46$ & $54.63$\\
2DMSE ($d=16$) & $67.79$ & $72.51$ & $69.61$ & $75.51$ & $71.74$ & $74.80$ & $73.09$ & $72.15$\\
\midrule
AnglE ($d=32$) & $31.84$ & $45.55$ & $38.22$ & $50.90$ & $55.63$ & $38.32$ & $53.19$ & $44.81$\\
MRL ($d=32$) & $51.57$ & $60.03$ & $52.50$ & $57.82$ & $64.91$ & $54.98$ & $58.44$ & $57.18$\\
2DMS ($d=32$) & $69.73$ & $77.13$ & $72.87$ & $77.94$ & $74.85$ & $77.84$ & $74.55$ & $74.99$\\
\midrule
AnglE ($d=64$) & $43.08$ & $49.07$ & $42.22$ & $56.56$ & $56.83$ & $46.79$ & $55.11$ & $49.95$\\
MRL ($d=64$) & $56.27$ & $61.88$ & $55.23$ & $62.65$ & $66.14$ & $59.41$ & $60.53$ & $60.30$\\
2DMSE ($d=64$) & $71.60$ & $79.33$ & $74.48$ & $80.00$ & $76.63$ & $79.35$ & $75.82$ & $76.74$\\
\midrule
AnglE ($d=128$) & $43.19$ & $51.14$ & $44.57$ & $57.80$ & $58.07$ & $48.17$ & $55.48$ & $51.20$\\
MRL ($d=128$) & $57.58$ & $63.10$ & $56.67$ & $65.58$ & $67.11$ & $61.29$ & $61.09$ & $61.77$\\
2DMSE ($d=128$) & $71.89$ & $80.28$ & $75.31$ & $81.06$ & $77.35$ & $80.58$ & $76.26$ & $77.53$\\
\midrule
AnglE ($d=256$) & $41.84$ & $52.02$ & $45.29$ & $57.55$ & $57.72$ & $47.88$ & $54.91$ & $51.03$\\
MRL ($d=256$) & $56.16$ & $63.13$ & $57.33$ & $65.62$ & $67.42$ & $61.87$ & $61.29$ & $61.83$\\
2DMSE ($d=256$) & $72.09$ & $80.91$ & $75.82$ & $81.92$ & $77.87$ & $81.34$ & $76.83$ & $78.11$\\
\midrule
AnglE ($d=512$) & $44.49$ & $46.94$ & $41.85$ & $56.22$ & $55.95$ & $46.54$ & $53.24$ & $49.32$\\
MRL ($d=512$) & $59.11$ & $60.71$ & $57.40$ & $67.57$ & $66.99$ & $63.45$ & $60.10$ & $62.19$\\
2DMSE ($d=512$) & $72.82$ & $80.32$ & $77.06$ & $82.99$ & $78.00$ & $81.90$ & $76.39$ & $78.50$\\
\midrule
AnglE ($d=768$) & $44.61$ & $47.61$ & $42.59$ & $57.31$ & $56.58$ & $46.96$ & $53.92$ & $49.94$\\
MRL ($d=768$) & $58.84$ & $61.52$ & $57.88$ & $68.06$ & $67.39$ & $63.83$ & $60.70$ & $62.60$\\
2DMSE ($d=768$) & $72.78$ & $80.83$ & $77.04$ & $83.04$ & $78.34$ & $81.96$ & $76.47$ & $78.64$\\
\midrule
\midrule
\multicolumn{9}{c}{$\#$ Layer $n=10$}\\
\midrule
AnglE ($d=8$) & $50.59$ & $50.23$ & $46.91$ & $56.53$ & $45.64$ & $47.25$ & $58.33$ & $50.78$\\
MRL ($d=8$) & $58.56$ & $62.83$ & $59.16$ & $59.58$ & $62.75$ & $63.10$ & $62.28$ & $61.18$\\
2DMSE ($d=8$) & $63.12$ & $67.89$ & $65.34$ & $70.73$ & $67.28$ & $69.94$ & $68.39$ & $67.53$\\
\midrule
AnglE ($d=16$) & $56.74$ & $57.25$ & $49.42$ & $59.66$ & $55.61$ & $55.47$ & $63.50$ & $56.81$\\
MRL ($d=16$) & $63.12$ & $68.44$ & $63.53$ & $65.26$ & $67.39$ & $66.53$ & $65.03$ & $65.61$\\
2DMSE ($d=16$) & $67.71$ & $75.09$ & $70.97$ & $75.99$ & $72.07$ & $75.04$ & $73.01$ & $72.84$\\
\midrule
AnglE ($d=32$) & $53.35$ & $63.09$ & $51.85$ & $61.06$ & $61.22$ & $57.70$ & $63.73$ & $58.86$\\
MRL ($d=32$) & $66.13$ & $73.22$ & $66.47$ & $69.28$ & $71.60$ & $70.78$ & $66.51$ & $69.14$\\
2DMSE ($d=32$) & $70.44$ & $79.56$ & $74.66$ & $78.50$ & $75.44$ & $78.60$ & $74.52$ & $75.96$\\
\midrule
AnglE ($d=64$) & $59.31$ & $65.90$ & $58.30$ & $66.09$ & $63.16$ & $64.85$ & $65.95$ & $63.37$\\
MRL ($d=64$) & $68.15$ & $75.91$ & $69.42$ & $72.24$ & $73.24$ & $73.94$ & $67.92$ & $71.55$\\
2DMSE ($d=64$) & $72.00$ & $81.70$ & $76.65$ & $80.58$ & $77.42$ & $80.47$ & $76.25$ & $77.87$\\
\midrule
AnglE ($d=128$) & $58.77$ & $67.59$ & $59.50$ & $66.81$ & $64.16$ & $66.37$ & $66.18$ & $64.20$\\
MRL ($d=128$) & $68.16$ & $77.85$ & $70.52$ & $73.68$ & $74.58$ & $75.65$ & $68.73$ & $72.74$\\
2DMSE ($d=128$) & $72.42$ & $82.82$ & $77.62$ & $81.68$ & $78.42$ & $81.89$ & $77.05$ & $78.84$\\
\midrule
AnglE ($d=256$) & $57.77$ & $67.45$ & $60.04$ & $66.92$ & $64.07$ & $66.28$ & $66.09$ & $64.09$\\
MRL ($d=256$) & $67.86$ & $78.18$ & $71.36$ & $74.78$ & $74.70$ & $76.62$ & $69.35$ & $73.26$\\
2DMSE ($d=256$) & $72.13$ & $83.32$ & $78.08$ & $82.40$ & $79.08$ & $82.63$ & $77.83$ & $79.35$\\
\midrule
AnglE ($d=512$) & $59.07$ & $63.21$ & $57.79$ & $66.32$ & $62.74$ & $64.23$ & $65.01$ & $62.62$\\
MRL ($d=512$) & $69.59$ & $76.81$ & $72.34$ & $75.98$ & $74.30$ & $76.43$ & $67.74$ & $73.31$\\
2DMSE ($d=512$) & $73.44$ & $83.47$ & $79.38$ & $83.68$ & $79.29$ & $83.13$ & $77.63$ & $80.00$\\
\midrule
AnglE ($d=768$) & $59.43$ & $64.16$ & $58.38$ & $67.23$ & $63.19$ & $64.82$ & $65.53$ & $63.25$\\
MRL ($d=768$) & $69.55$ & $77.57$ & $72.54$ & $76.55$ & $74.63$ & $76.88$ & $68.37$ & $73.73$\\
2DMSE ($d=768$) & $73.33$ & $83.85$ & $79.31$ & $83.70$ & $79.52$ & $83.30$ & $77.83$ & $80.12$\\
\midrule
\midrule
\multicolumn{9}{c}{$\#$ Layer $n=11$}\\
\midrule
AnglE ($d=8$) & $55.75$ & $62.27$ & $56.75$ & $62.93$ & $56.41$ & $57.73$ & $64.89$ & $59.53$\\
MRL ($d=8$) & $63.51$ & $69.66$ & $65.51$ & $67.15$ & $67.22$ & $70.84$ & $66.41$ & $67.19$\\
2DMSE ($d=8$) & $63.51$ & $69.30$ & $65.99$ & $69.63$ & $67.73$ & $70.02$ & $68.64$ & $67.83$\\
\midrule
AnglE ($d=16$) & $65.21$ & $67.76$ & $61.99$ & $69.40$ & $65.90$ & $68.81$ & $70.69$ & $67.11$\\
MRL ($d=16$) & $68.38$ & $76.16$ & $71.63$ & $74.40$ & $73.04$ & $76.45$ & $70.24$ & $72.90$\\
2DMSE ($d=16$) & $68.57$ & $76.40$ & $72.13$ & $76.04$ & $73.23$ & $75.65$ & $73.23$ & $73.61$\\
\midrule
AnglE ($d=32$) & $69.74$ & $74.26$ & $67.83$ & $73.26$ & $73.12$ & $75.84$ & $75.39$ & $72.78$\\
MRL ($d=32$) & $71.77$ & $80.72$ & $75.25$ & $78.82$ & $77.03$ & $80.18$ & $72.59$ & $76.62$\\
2DMSE ($d=32$) & $71.51$ & $81.37$ & $76.21$ & $79.12$ & $77.09$ & $79.76$ & $75.39$ & $77.21$\\
\midrule
AnglE ($d=64$) & $71.70$ & $77.12$ & $71.90$ & $76.61$ & $75.53$ & $78.57$ & $77.08$ & $75.50$\\
MRL ($d=64$) & $73.51$ & $82.87$ & $77.59$ & $80.96$ & $79.02$ & $81.67$ & $74.44$ & $78.58$\\
2DMSE ($d=64$) & $73.32$ & $83.57$ & $78.45$ & $81.41$ & $79.21$ & $81.90$ & $77.28$ & $79.31$\\
\midrule
AnglE ($d=128$) & $70.75$ & $78.68$ & $73.00$ & $78.64$ & $76.62$ & $80.09$ & $77.54$ & $76.47$\\
MRL ($d=128$) & $73.25$ & $84.11$ & $78.43$ & $82.31$ & $79.97$ & $82.91$ & $75.63$ & $79.52$\\
2DMSE ($d=128$) & $73.53$ & $84.61$ & $79.39$ & $82.84$ & $80.05$ & $83.11$ & $78.26$ & $80.26$\\
\midrule
AnglE ($d=256$) & $70.94$ & $78.83$ & $73.81$ & $80.52$ & $76.60$ & $80.55$ & $77.46$ & $76.96$\\
MRL ($d=256$) & $73.03$ & $84.26$ & $79.07$ & $83.68$ & $80.42$ & $83.54$ & $76.46$ & $80.07$\\
2DMSE ($d=256$) & $73.11$ & $84.80$ & $79.85$ & $84.00$ & $80.75$ & $83.71$ & $79.07$ & $80.76$\\
\midrule
AnglE ($d=512$) & $72.10$ & $74.66$ & $73.83$ & $80.96$ & $75.95$ & $80.65$ & $75.34$ & $76.21$\\
MRL ($d=512$) & $74.57$ & $83.17$ & $80.41$ & $84.11$ & $80.39$ & $83.41$ & $74.45$ & $80.07$\\
2DMSE ($d=512$) & $74.50$ & $84.83$ & $81.36$ & $85.36$ & $80.83$ & $84.26$ & $78.32$ & $81.35$\\
\midrule
AnglE ($d=768$) & $72.73$ & $76.37$ & $74.39$ & $81.70$ & $76.51$ & $81.09$ & $76.13$ & $76.99$\\
MRL ($d=768$) & $74.70$ & $84.12$ & $80.66$ & $84.72$ & $80.70$ & $83.75$ & $75.36$ & $80.57$\\
2DMSE ($d=768$) & $74.32$ & $85.39$ & $81.39$ & $85.49$ & $81.04$ & $84.48$ & $78.77$ & $81.55$\\
\midrule
\midrule
\multicolumn{9}{c}{$\#$ Layer $n=12$}\\
\midrule
AnglE ($d=8$) & $57.45$ & $67.73$ & $60.77$ & $67.17$ & $60.64$ & $62.19$ & $65.98$ & $63.13$\\
MRL ($d=8$) & $64.42$ & $71.68$ & $67.66$ & $70.87$ & $68.60$ & $72.26$ & $68.29$ & $69.11$\\
2DMSE ($d=8$) & $63.04$ & $69.74$ & $67.01$ & $72.14$ & $68.42$ & $70.69$ & $69.49$ & $68.65$\\
\midrule
AnglE ($d=16$) & $65.03$ & $75.86$ & $68.69$ & $73.76$ & $70.58$ & $72.81$ & $72.37$ & $71.30$\\
MRL ($d=16$) & $70.48$ & $79.16$ & $74.10$ & $78.55$ & $74.84$ & $78.99$ & $73.42$ & $75.65$\\
2DMSE ($d=16$) & $69.60$ & $77.91$ & $74.29$ & $78.88$ & $74.92$ & $77.37$ & $75.22$ & $75.46$\\
\midrule
AnglE ($d=32$) & $71.87$ & $80.53$ & $73.55$ & $77.23$ & $77.16$ & $79.44$ & $76.11$ & $76.56$\\
MRL ($d=32$) & $73.63$ & $83.42$ & $77.70$ & $82.55$ & $78.00$ & $82.23$ & $75.92$ & $79.06$\\
2DMSE ($d=32$) & $72.63$ & $83.11$ & $78.49$ & $82.26$ & $78.62$ & $81.49$ & $77.48$ & $79.15$\\
\midrule
AnglE ($d=64$) & $73.20$ & $82.90$ & $77.08$ & $81.16$ & $80.42$ & $82.30$ & $79.06$ & $79.45$\\
MRL ($d=64$) & $74.89$ & $85.06$ & $79.65$ & $84.19$ & $80.13$ & $83.58$ & $77.72$ & $80.75$\\
2DMSE ($d=64$) & $74.35$ & $84.99$ & $80.45$ & $84.18$ & $80.67$ & $83.74$ & $78.99$ & $81.05$\\
\midrule
AnglE ($d=128$) & $74.10$ & $84.18$ & $78.52$ & $83.13$ & $81.56$ & $84.12$ & $80.36$ & $80.85$\\
MRL ($d=128$) & $75.29$ & $85.97$ & $80.54$ & $85.57$ & $80.94$ & $84.51$ & $78.41$ & $81.60$\\
2DMSE ($d=128$) & $74.68$ & $85.88$ & $81.22$ & $85.53$ & $81.51$ & $84.76$ & $79.67$ & $81.89$\\
\midrule
AnglE ($d=256$) & $74.17$ & $84.98$ & $79.38$ & $85.07$ & $81.89$ & $84.90$ & $80.85$ & $81.61$\\
MRL ($d=256$) & $75.08$ & $86.20$ & $81.07$ & $86.34$ & $81.30$ & $84.84$ & $79.00$ & $81.98$\\
2DMSE ($d=256$) & $74.52$ & $86.17$ & $81.72$ & $86.06$ & $81.93$ & $85.21$ & $79.97$ & $82.23$\\
\midrule
AnglE ($d=512$) & $75.12$ & $84.86$ & $80.50$ & $86.23$ & $82.44$ & $85.76$ & $80.72$ & $82.23$\\
MRL ($d=512$) & $75.90$ & $86.57$ & $81.86$ & $86.72$ & $81.72$ & $85.57$ & $79.17$ & $82.50$\\
2DMSE ($d=512$) & $75.09$ & $86.49$ & $82.29$ & $86.46$ & $82.02$ & $85.73$ & $80.04$ & $82.59$\\
\midrule
AnglE ($d=768$) & $75.26$ & $85.61$ & $80.64$ & $86.36$ & $82.51$ & $85.64$ & $80.99$ & $82.43$\\
MRL ($d=768$) & $75.72$ & $86.79$ & $81.89$ & $86.91$ & $81.74$ & $85.50$ & $79.44$ & $82.57$\\
2DMSE ($d=768$) & $75.00$ & $86.69$ & $82.30$ & $86.50$ & $82.09$ & $85.79$ & $80.18$ & $82.65$\\
\bottomrule
\caption{Detailed STS Benchmark results of scalable sentence embeddings. BERT$_{base}$ serves as the backbone for all models.}
\label{table_detailed_results}
\end{longtable}
\end{document}